\documentclass{article}

\usepackage{acra}
\usepackage{graphicx}
\usepackage{multirow}
\usepackage[table,xcdraw]{xcolor}
\usepackage{adjustbox}
\usepackage{amsmath}
\usepackage[hyphens]{url}
\usepackage{subcaption}
\usepackage{hyperref}
\usepackage{siunitx}
\usepackage[labelfont=bf]{caption}
\usepackage{color, colortbl} 
\definecolor{Gray}{gray}{0.9} 

\title{OrchardDepth: Precise Metric Depth Estimation of Orchard Scene from Monocular Camera Images}
\author{
  Zhichao Zheng$^*$, Henry Williams,  Bruce A MacDonald$^{**}$\\
  Centre for Automation and Robotic Engineering Science\\
  The University of Auckland, NZ\\
  \texttt{zzhe013@aucklanduni.ac.nz}$^*$, \texttt{b.macdonald@auckland.ac.nz}$^{**}$ \\
}

\begin{document}

\maketitle
\begin{abstract}
    Monocular depth estimation is a rudimentary task in robotic perception. Recently, with the development of more accurate and robust neural network models and different types of datasets, monocular depth estimation has significantly improved performance and efficiency. However, most of the research in this area focuses on very concentrated domains. In particular, most of the benchmarks in outdoor scenarios belong to urban environments for the improvement of autonomous driving devices, and these benchmarks have a massive disparity with the orchard/vineyard environment, which is hardly helpful for research in the primary industry. Therefore, we propose OrchardDepth, which fills the gap in the estimation of the metric depth of the monocular camera in the orchard/vineyard environment. In addition, we present a new retraining method to improve the training result by monitoring the consistent regularization between dense depth maps and sparse points. Our method improves the RMSE of depth estimation in the orchard environment from 1.5337 to 0.6738, proving our method's validation.
\end{abstract}

\section{Introduction}
    Depth is essential information in mobile robotic systems. Systems such as 6DoF pose estimation \cite{li2023depth,bundlesdfwen2023}, autonomous driving \cite{xue2020toward,cheng2024adaptive}, and visual SLAM \cite{zhu2020camvox,davison2007monoslam} rely on the accuracy of depth data to function. Conventional methods based on depth sensors face limitations such as sparseness, low resolution, and pattern mismatch. In addition, data from different types of sensors with different resolutions and fields of view (FOV) require additional calibration to solve the disparity between stationary and moving observations and the challenge of time synchronization between sensors. Therefore, comprehensive data with visual and depth information from the same sensor would significantly reduce the mismatch between these data and help improve the performance of downstream tasks such as obstacle detection and collision avoidance. 
    
    \begin{figure}[h]
        \centering
        \includegraphics[width=1\linewidth]{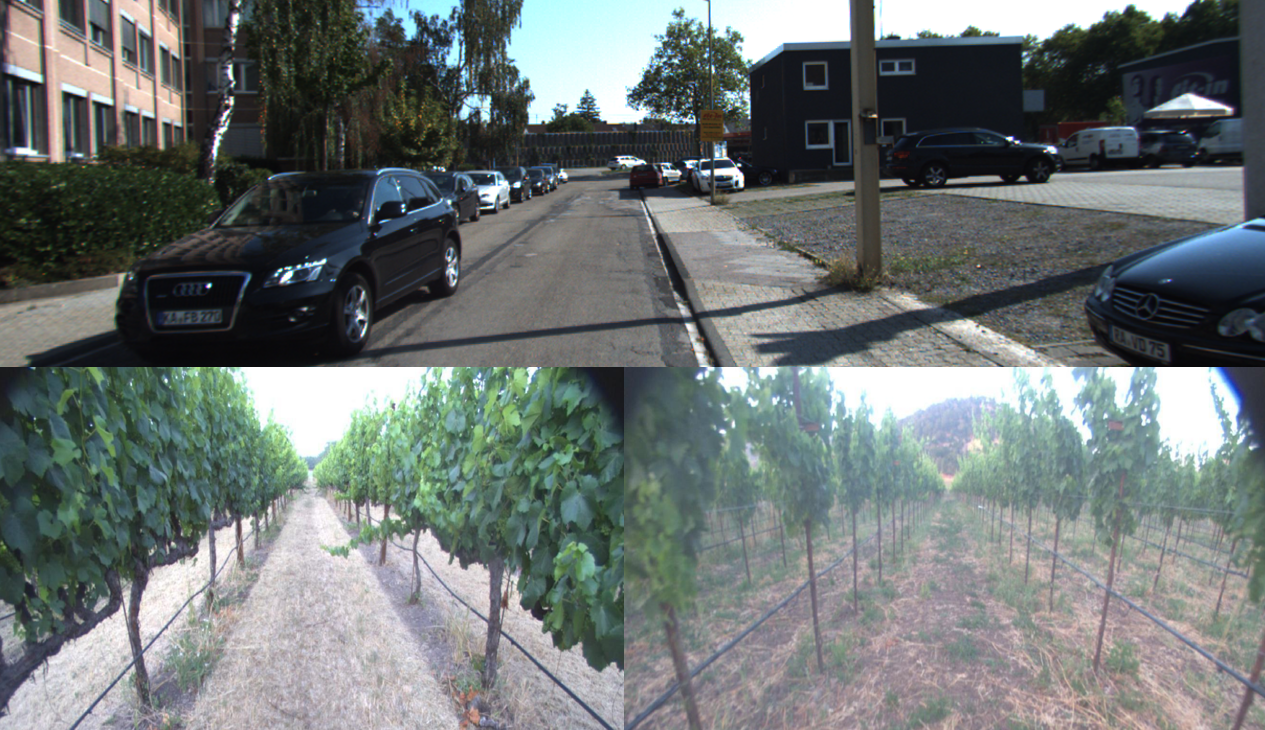}
        \caption{\textbf{Scene Display} - Illustrates the disparity between the orchard and the city scene. (\textbf{Top}) Image sample from KITTI depth dataset. (\textbf{Bottom}) Image captured in an Orchard from the US.}
        \label{fig:Scene Display}
    \end{figure}
    
    Recently, most studies have focused on high-resolution and high-precision monocular depth estimation (MDE) models \cite{li2023patchfusion,depth_anything_v1,depth_anything_v2}, which allow the acquisition of complete RGB-D data from a single RGB image captured by a monocular camera. Furthermore, the prediction performance of the state-of-the-art MDE models, enabled by the development of fundamental learning-based algorithms, represents a massive improvement and achieves zero-shot capabilities in 2D metric benchmarks.

    Recent studies in this area have focused on indoor scenes \cite{nyudepthv2} or outdoor scenes \cite{Cordts2016Cityscapes,kitti,waymo,nuscenes} based on urban streets. None provide research data and benchmarks in rural areas, especially in orchards and vineyards. Meanwhile, the encode-decode architecture has been widely used in MDE tasks, and the encode module needs to be trained on extensive and varied datasets to maintain resilience. Therefore, due to the significant environmental differences, the SOTA models trained in urban scenes will lose their zero-shot ability and show a significant drop in performance in rural areas. In this paper, we build our depth estimation model to address this issue based on the image and sparse LiDAR point cloud image from orchard/vineyard environments in New Zealand, Australia, and the United States. 

    In addition, we have collected a large amount of data in the orchard using a monocular camera and LiDAR, which contains only RGB images and sparse point clouds. Previous research has shown that training a depth estimation model solely on sparse points can lead to performance degradation due to the limited depth information and the lack of continuity and connectivity between objects within the field of view. Therefore, inspired by semi-supervised training methods \cite{crossconsistency}, we introduced a novel method to supervise the consistency between the dense depth map generated by the stereo camera and the projected LiDAR point ground truth in the KITTI \cite{kittdepthbenchmard} dataset, helping the custom dataset to maintain dense consistency while training on the sparse depth map.

    In summary, we have contributed to the following in this study:
    \begin{itemize}
        \item A novel depth estimation model trained on the vineyard and orchard scenes fills the gap of the MDE model in the rural environment.
        \item Our study proposes a new method that monitors the consistency between the dense depth map and the sparse points of existing datasets to take advantage of the training with the data captured in the orchard with only sparse points.
        \item We achieve the SOTA performance in the MDE tasks in the orchard environment and enlighten the method for the studies in related fields in the primary industry.
    \end{itemize}
    
\section{Background and Related Work}
Recent studies can be divided into two main groups: affine invariant inverse depth estimation and metric depth estimation. Affine invariant inverse depth aims to estimate the disparity depth between objects. Metric depth estimation focuses on outputting depths in standard units such as meters or millimeters.
\subsection{Affine-Invariant Inverse Depth Estimation}
The learning-based method typically uses large and varied datasets to eliminate bias and reduce the degradation of the model, thus maintaining the robustness of the model. For example, camera-specific intrinsic parameters can cause a mismatch between feature and depth information, which blocks the convergence of the model. Therefore, instead of directly estimating depth in meters, some studies predict the relative depth disparity between objects to avoid the scale and shift offset caused by focal length and distortion. MiDaS\cite{MiDaS} pioneered a zero-shot cross-dataset transfer protocol by training its network on multiple datasets and validating its results across the different datasets. MiDaS also introduced scale and shift-invariant loss, which helps the depth estimation models to converge across different datasets. The following version, MiDaSv3.1\cite{MiDaSv3.1}, uses multiple vision transformer-based encoders \cite{beit,dosovitskiy2020vit,liu2021Swin,liu2021swinv2} and dense prediction transformer decoders \cite{dpt} to provide better performance and runtime options for downstream tasks, with the largest model in this framework improving the quality of depth estimation by over 28\%.

More recently, DepthAnything\cite{depth_anything_v1} provided a new foundation model to achieve a more robust zero-shot depth prediction capability than Midasv3.1 by using the unlabelled images to enrich the train data, thus reducing the generalization error. They used a powerful teacher model to generate pseudo-depth annotations for unlabelled images and frozen weights of DINOv2\cite{oquab2023dinov2} as the encoder. They also introduced a feature alignment loss to preserve the semantic prior from the encoder and maintain the model's ability to understand the unseen scene.

\cite{Marigold} reports that scene understanding is essential for depth estimators to predict depth based on scene content accurately. Therefore, a stable diffusion model that can provide comprehensive and encyclopedic prior knowledge would significantly improve the performance and cross-domain capability of the depth estimator. Hence, \cite{Marigold} introduces a latent diffusion model called Marigold, which is fine-tuned by the denoising U-net with synthetic RGB-D data. This approach achieves advanced performance on various natural images. Furthermore, the network was trained on synthetic data and achieved zero-shot transfer to real data. This makes the method more robust to edge cases wrongly acquired by depth sensors, such as transparent and reflective objects.

Compared to Marigold, the Depth Anything model is more sensitive to reflective surfaces and transparent objects, and the prediction of object edges needs to be smoother. To address this, Depth Anything v2 \cite{depth_anything_v2} provides a more resilient model that is more robust in complex scenes. It achieves finer details, surpassing both the Marigold and DepthAnything. The essential method to solve this problem is data. DepthAnything v2 trained its DINOv2-G-based teacher model only on synthetic images, which can avoid the noise from the sensor depth label, which misleads the result by failing to detect the transparent and reflective surfaces. This study also introduces a method of using real images as intermediate learning targets to eliminate the distribution shift problem for direct training with synthetic images.

In general, recent studies in relative depth estimation achieved significant improvement. The model in this field provides fine-grained details, highly accurate results, and zero-shot capability in unseen scenes. However, most real-world scenes need to precisely predict metric depth in meters, which can directly implement the result to the application.
    
\subsection{Monocular Metric Depth Estimation}
Recent research shows significant differences between indoor and outdoor environments regarding space density and depth boundary. Thus, the distribution is critical in helping us sort it during depth regression. DORN \cite{FuCVPR18-DORN} explores the relationship between depth distribution and regression accuracy. It reports that, in distant scenes, features extracted by a Conv-based neural network contribute less helpful information. Instead, these features can introduce noise into the depth prediction.
To address this, \cite{FuCVPR18-DORN} suggests dividing the depth range in a scene into five subintervals by space-increasing discretization. The model also down-weights the loss for the distant scene to maintain the precise nearby. Hence, the prediction can accurately handle different depth distributions at various distances.

Based on DORN, AdaBins\cite{bhat2021adabins} report that the depth distribution represents massive differences across different images, and apparently, sharp depth discontinuities learned from one image can severely affect the result in others. Therefore, AdaBins denotes the depth subintervals as bins, and the bin widths represent the depth range distribution learned from a transformer-based encoder. Since the distribution changes with the images, the model uses an MLP head to predict the correct bin width based on the prior knowledge learned by the encoder. The bin width then becomes the depth prediction in a few intervals. The final depth prediction can be calculated with the linear combination of the bin center.

LocalBins\cite{bhat2022localbins} reports that AdaBins global adaptive bins heavily rely on the transformer-based architecture and features extracted near the output layer of the model. Otherwise, the network exhibits unstable training divergence and convergence at a local minimum, known as the ``late injection problem". To avoid the drawback of global adaptive bin prediction, LocalBins attempts to estimate a per-pixel bin partition corresponding to the depth distribution in adjacent local pixels. Unlike AdaBins, LocalBins uses the bottleneck and decoder functions of the encoder-decoder network, which can learn better depth representations through distribution monitoring.

ZoeDepth\cite{bhat2023zoedepth} introduces a two-stage framework that further improves the performance of metric depth estimation by extracting prior knowledge from relative depth estimation. The model improves the architecture of MidiasV3.1 by replacing the backbone with $BEiT_{384}$-L\cite{beit} to get better performance on relative depth prediction in the first stage, and training the model with extensive and varied depth data to maintain the generalization of the model. The feature extraction from multiple layers in the encoder is then fed into the backbone and fusion in the Metric Bins module\footnote{Introduce an inverse attractor to adjust the bin center rather than split the bin.} to produce the metric depth prediction. This framework was flexible enough to be modified with the encoder-decoder relative depth estimation approaches. The first stage is compatible with the ViT-encoder structure, thus DepthAnything and DepthAnythingV2 can be used as the first stage relative depth predictor, and the metric depth can be fine-tuned using the Metric Bins module of ZoeDepth.

Although ZoeDepth achieves metric depth estimation through a two-step process, this capability is obtained by fine-tuning the dataset of different scenes, such as NYUv2 or KITTI, which is still strongly related to the camera-specific intrinsic parameters. Unlike two-stage methods, UniDepth\cite{piccinelli2024unidepth} tries to predict metric depth directly from an arbitrary image captured in any scene. They show that decoupling the camera and depth representations is the key to solving this problem. Specifically, they transform the 3D point representation from the orthogonal base Cartesian coordinate (x, y, z) to a pseudo-spherical coordinate with azimuth, elevation, and log-depth ($\theta, \phi, logz$) through a self-promoting camera module based on a transformer structure, thus decoupling the angles of the pinhole camera ray from depth in a natural way. To ensure consistency, they then implement a geometric invariance loss to monitor the camera-prompted depth feature of the same scene captured by different cameras.

Meanwhile, Metric3D\cite{yin2023metric} uses a more straightforward method to resolve the metric ambiguity caused by sensor disparity in different datasets. By empirically studying how sensor settings such as pixel size, focal length, and sensor size affect depth prediction, they found that focal length is the critical factor affecting the result. Therefore, to eliminate the metric ambiguity of the focal length, Metric3D introduces a novel transformation method to transform either the image appearance or the ground truth label into a canonical camera space and supervises the training with a randomly proposed normalization loss, which can not only decouple the scale difference but also emphasize the local geometry and depth distribution from a single image. The subsequent work, Metric3Dv2 \cite{hu2024metric3dv2}, simultaneously predicts the metric depth and the surface normal, as they are highly complementary since the metric depth provides large training datasets and the surface normal contains rich local geometry information that can give geometrical constraints on the metric depth during training.

In summary, previous metric depth estimation methods have significantly improved both the zero-shot capabilities of in-the-wild images and the prediction accuracy. However, most of these studies were based on data from public urban roads or indoor scenes. In addition, some attempted to solve extreme cases, such as reflective surfaces or transparent objects, but none were based on images of primary industry scenes such as vineyards or orchards. The lack of scene-based prior knowledge will eventually degrade the performance of these models in an orchard. Therefore, we propose OrchardDepth to predict the depth of the orchard in a more accurate environment.

    \begin{figure}[htbp]
         \centering
         \begin{subfigure}[t]{0.23\textwidth}
             \centering
             \includegraphics[width=\textwidth]{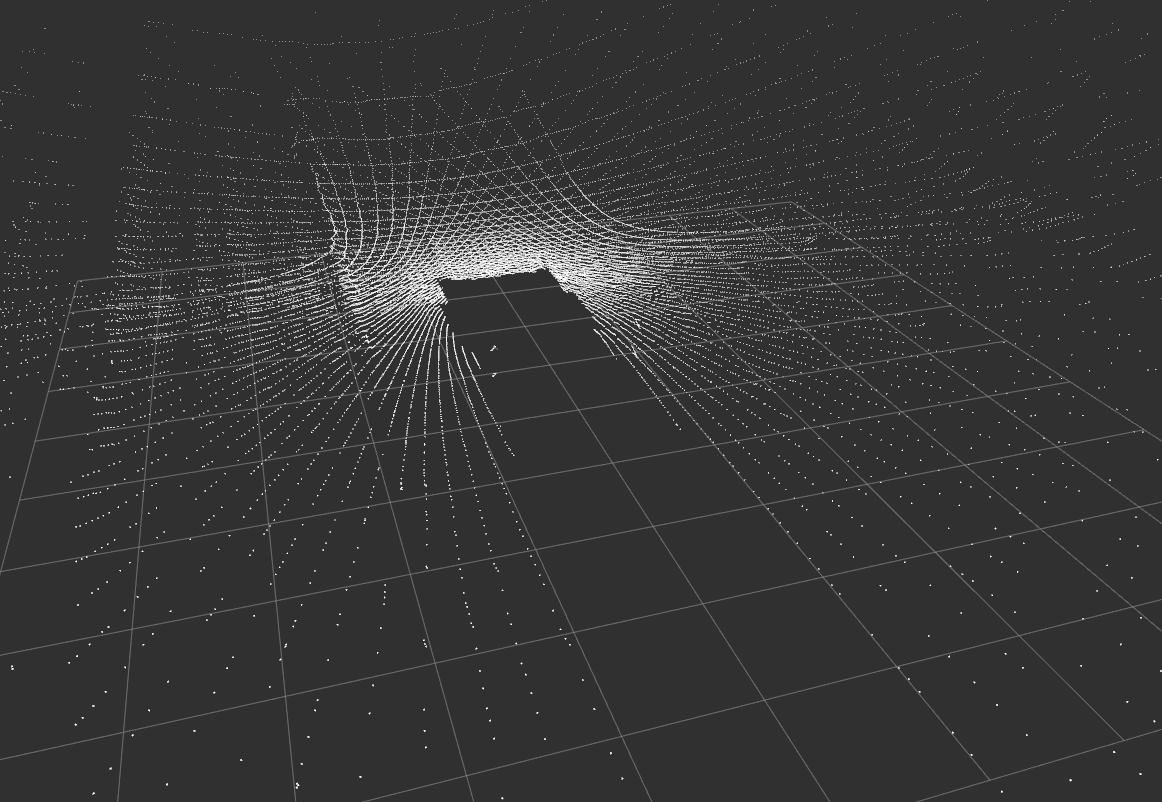}
             \caption{Combined LiDAR Points}
             \label{fig:rviz lidar points}
         \end{subfigure}
         \hfill
         \begin{subfigure}[t]{0.23\textwidth}
             \centering
             \includegraphics[width=\textwidth]{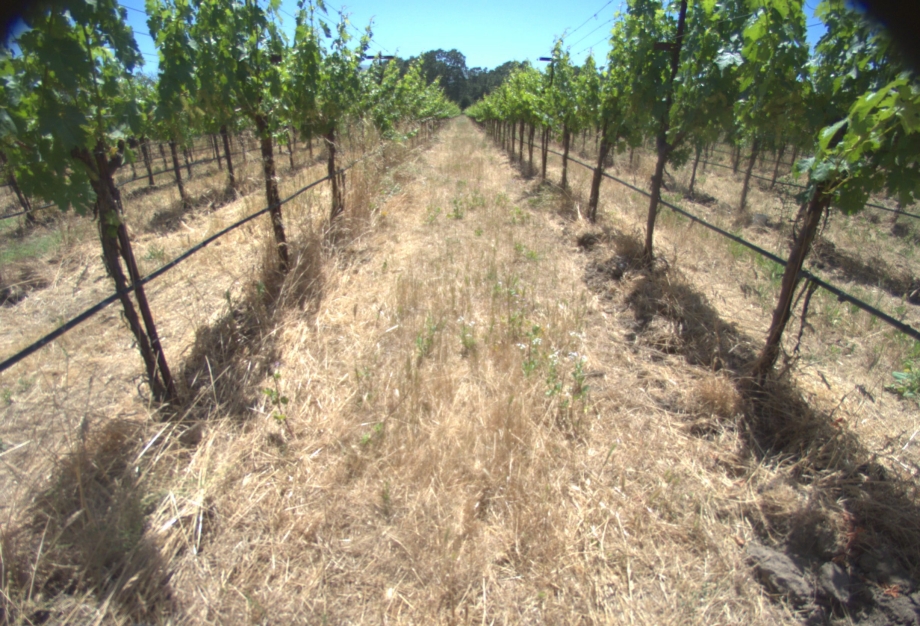}
             \caption{RGB Camera Image}
             \label{fig:orchard scene 1}
         \end{subfigure}
         \hfill
         \begin{subfigure}[t]{0.23\textwidth}
             \centering
             \includegraphics[width=\textwidth]{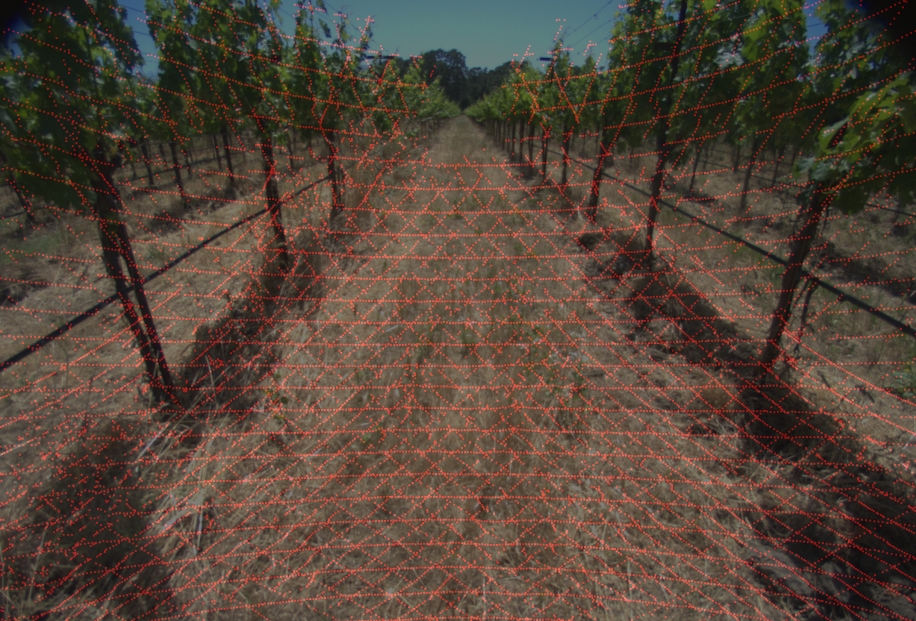}
             \caption{Projected Points}
             \label{fig:visual points in orchard 1}
         \end{subfigure}
            \caption{ \textbf{Data Preparation} - An illustration of the calibration and projection of different sensors. \textbf{(a)} Combined LiDAR Points acquisition from three LiDAR sensors. \textbf{(b)} Image captured from the center camera. \textbf{(c)} Projected Combined LiDAR points to camera space}
            \label{fig:Data Preparation Demostration}
    \end{figure}
    
\begin{figure*}[t]
    \centering
    \includegraphics[width=\textwidth]{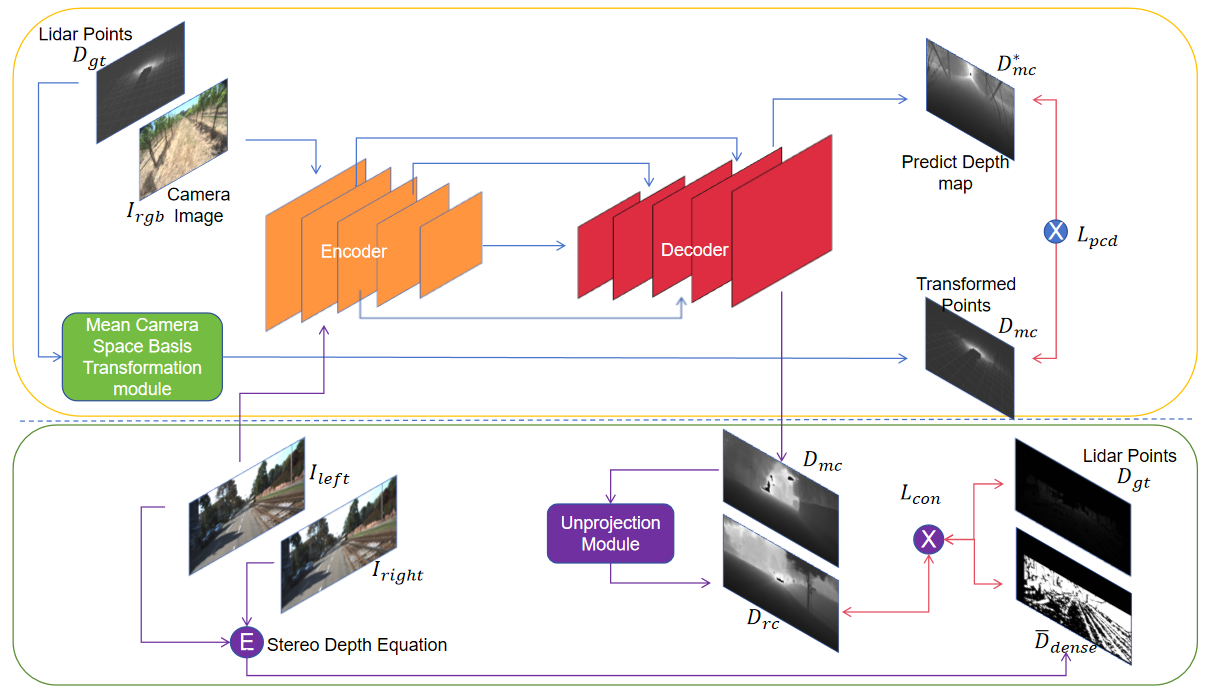}
    \caption{\textbf{Pipeline} - \textit{Top}: The training process with the custom dataset, the depth ground truth is obtained from the combined points from three LiDAR sensors, and the points acquired from LiDAR sensors will be projected to the camera coordinate system and then transformed into the mean camera space. \textit{Bottom}: In the training process with the KITTI dataset, we first generate dense depth with the stereo camera image. Then, we use the left image as the input to predict the depth map and convert it back to the acquisition camera coordinate, calculate the loss of predicted depth between dense depth and KITTI ground truth points, respectively, and then use \(L_{con}\) to supervise the dense-sparse consistency.}
    \label{fig:Model Architecture}
\end{figure*}

\section{Methods}\label{sec:methods}
    \textbf{Data Preparation.} Our training data is derived from two primary sources: the KITTI depth\cite{kittdepthbenchmard} and self-collected data from an RGB camera, and three 32-line combined LiDAR points. We first calibrated the LiDAR-LiDAR and LiDAR-camera coordinates using a conventional checkerboard method to generate relatively dense depth information through sparse points. Next, we combined the LiDAR points acquired by the LiDARs and then projected the combined points onto the RGB camera coordinate. Then, we used these projected points as supervision to train the depth network. Meanwhile, for the raw KITTI dataset \cite{rawkitti}, we use conventional stereo camera depth estimation by using the rectified image acquisition from the left and right cameras, and calculate the dense depth map with the following equation.

    \begin{equation}
        D_{dense} = \frac{f \times B} {\bar{dp}}
    \end{equation}

    Where \(D_{dense}\) represents the dense depth map generated by the stereo camera, \(f\) is the focal length, B and \(\bar{dp}\) are the baseline and disparity of the stereo camera, respectively. In this paper, we use \(D_{dense}\), and KITTI projected LiDAR points to supervise the depth map generated from the image captured by the left camera.

\subsection{Architecture}
\textbf{Mean Camera Space Basis Transform Module.} Predicting a dense depth map from a single RGB image is a well-known ill-posed problem since the correspondence between the feature and depth of an object in the camera view requires a distance constraint. This constraint is usually provided by the camera's intrinsic parameters, in particular the focal length. Inspired by \cite{yin2023metric} and \cite{hu2024metric3dv2}, we will first map the RGB image into a canonical camera space. However, unlike \cite{yin2023metric,hu2024metric3dv2}, which creates an arbitrary canonical camera space that is not based on the sensor parameters of all datasets, we create the closest canonical camera space to the focal lengths of all data collecting sensors \( \{f_i\}_{i=1}^n \) by calculating their mean focal length \(\bar{f}_{\text{mc}}\). 

\begin{equation}
    \bar{f}_{\text{mc}} = \frac{1}{N} \sum_{i=1}^{n} f_i
\end{equation}

Since our camera intrinsic settings are close enough, it will essentially help to reduce the accuracy loss from the coordinate base transformation of the custom dataset containing only sparse LiDAR points.

Next, we project the sparse depth point into the mean canonical camera space.

\begin{equation}
    {D}_{mc} = \frac{\bar{f}_{mc}}{f_{gt}} \times D_{gt}
\end{equation}

Where \(D_{mc}\) , \(D_{gt}\) and \(f_{gt}\) are the depth labels in mean canonical camera space, ground truth depth label and ground truth camera focal length respectively.

\vspace{10pt}
\noindent \textbf{Encoder-Decoder Module.} Encoder-decoder architectures have proven to be highly efficient for dense prediction tasks such as depth estimation. The encoder extracts features from the large training data, and the decoder reconstructs the specific information through representations of the multi-scale features passed by the encoder. Recently, the encoder based on the vision transformer \cite{darcet2023vitneedreg} has been trained with large amounts of data from different scenes, which is already sufficient to retrieve the feature representation in orchard/vineyard scenes. In this paper, we focus on the decoding part, which reconstructs the depth of information from the features. Inspired by the dense prediction transformer \cite{dpt}, we use the DPT header as our decoder to build the correspondence between the feature and depth information in the orchard/vineyard scene. For this reason, we freeze the coding with DINOv2 \cite{oquab2023dinov2} pre-trained weights and train the decoder with the combined data from the orchards and the KITTI depth. 

\subsection{Supervision}
\textbf{Sparse Points Supervision.} In order to train the model with the combined LiDAR points collected from various orchards, two directions for applying the loss function to supervise the learning process need to be considered: cross-data consistency and the presence of outlier points resulting from the limited number of occlusion points. Inspired by MidaS \cite{MiDaS}, we chose the scale-invariant loss to supervise the custom dataset.

\begin{multline}
L_{\text{silog}} = \frac{1}{2|\mathbf{N^+}|} \sum_{i=1}^{|\mathbf{N^+}|} \left( \log D_{mc}^+ - \log D_{mc}^{*+} \right)^2 - \\
\frac{\lambda}{2|\mathbf{N^+}|^2} \left( \sum_{i=1}^{|\mathbf{N^+}|} \left( \log D_{mc}^+ - \log D_{mc}^{*+} \right) \right)^2
\end{multline}

Where \(N^+\) represents the number of positive depth points in the ground truth depth map, \(D_{mc}^{+}\) and \(D_{mc}^{*+}\) represent the predicted depth and ground truth depth in mean camera space respectively. The first term of this equation is an MSE loss that penalizes the outlier depth points, and the second term is an element-wise L1 loss that ensures data consistency across sensors. A weighting ratio \(\lambda\) has been used to balance these terms, and we use \(\lambda\) = 0.3 by empirical study in this paper.

\vspace{10pt}
\noindent \textbf{Dense-Sparse Consistency Supervision.} Training the MDE model solely on the sparse depth points acquired by the LiDAR sensors is constrained by many terms that we cannot eliminate in our custom datasets. First, the accuracy of the combined points is primarily related to the accuracy of the calibration, which is inherently less accurate than the data acquired by the unique sensor. Secondly, there will be some points in the combined point cloud that are occluded in the FOV of the camera, which is caused by the side-mounted LiDARs, which may project onto the same pixel of the image as the object that has blocked the view in the center, causing the ambiguity of depth. Finally, the combined points contained only discontinuous depth points, which may not accurately represent the distribution at the edge of an object due to the lack of information in between. Therefore, we use the KITTI dataset \cite{kittdepthbenchmard,kitti} to generate a dense depth map and also use its sparse point, which is an auxiliary supervision, to check the sparse-dense consistency, which ensures that supervising with sparse points can lead our model to generate a depth close enough to the model supervised by the depth map.

In particular, we first recover the predicted depth map from the mean camera space with \(D_{rc} = ({f_{gt}} / {\bar{f}_{mc}}) \times D_{mc}\) to the acquisition camera space to align the depth ground truth. 

Next, we calculate the SiLog loss between the predicted depth in camera coordinates \(D_{rc}\) with points \(D_{gt}\) and the dense depth map \(\bar{D}_{dense}\) ground truth. They are then normalized using the following equation:

\begin{equation}
    L_{silog}^{norm}(X, Y) = \frac{L_{silog}(X)}{L_{silog}(X) + L_{silog}(Y) + 1e^{-6}}
\end{equation}

Then we calculate the prediction consistency with sparse and dense depth maps by their SiLog norm using the following equation:

\begin{equation}
    L_{con} = MSE((L_{silog}^{norm}(L_{gt}, L_{dense}) 
    , L_{silog}^{norm}(L_{dense}, L_{gt}))
\end{equation}

Where \(L_{gt}\) and \(L_{dense}\) represent the SiLog loss between predicted depth with sparse LiDAR points and dense stereo depth respectively. 

Finally, the final loss function takes into account the discrepancy between the predicted depth with sparse points, dense depth and consistency, as represented by the following equation.

\begin{equation}
    L_{Final} = \alpha L_{silog}(D_{rc}, D_{gt}) + \beta  L_{silog}(D_{rc}, \bar{D}_{dense}) + \gamma L_{con}
\end{equation}

Where \(\alpha\), \(\beta\) and \(\gamma\) are learnable parameters to control the contributions of each loss component. Where \(\alpha\), \(\beta\) start with a value of 1.2 and clamp in the range between (0, 2.0), and \(\gamma\) starts with a value of 0.5 and clamp in the range between (0, 1.0) during the training process.

    \begin{table*}[ht]
        \centering
        \begin{adjustbox}{width=\textwidth}
            \begin{tabular}{c|c|ccc|c|c|c|c}
            \hline
             \multicolumn{2}{c}{} & \multicolumn{3}{|c|}{Higher Better} & \multicolumn{4}{c}{Lower Better} \\ 
            \hline
            Validation Data  & Loss & $\delta_1$ & $\delta_2$ & $\delta_3$ & AbsRel & RMSE & $RMSE\_log$ & log10 \\
            \hline
       
               \multirow{2}{*}{KITTI\_{proj\_points}}  
                & SiLog                  & 0.8565 & 0.9612 & 0.9883 & 0.1124 & 5.1992  & 0.1746 & 0.0526 \\
                & ConsistencyLoss(Ours)   & \textbf{0.8566↑} & 0.9492 & 0.9822 & \textbf{0.1068↓} & \textbf{2.8063↓}  & 0.1865 & \textbf{0.0495↓} \\
            \hline
    
                 \multirow{2}{*}{KITTI\_{stereo\_depth}} 
                & SiLog                  & 0.6335 & 0.8351 & 0.9385 & 0.2546 & 10.3845 & 0.3245 & 0.09996 \\
                & ConsistencyLoss(Ours)   & \textbf{0.8043↑} & \textbf{0.9449↑} & \textbf{0.9796↑} & \textbf{0.1472↓} & \textbf{2.3461↓}  & \textbf{0.2207↓} & \textbf{0.0632↓} \\
            \hline

                \multirow{2}{*}{Custom\_Orchard} 
                & SiLog                  & 0.9219 & 0.9753 & 0.98983 & 0.08599 & 1.5337 & 0.1421 & 0.0372 \\
                & ConsistencyLoss(Ours)   & 0.9231↑ & 0.9746 & 0.9897 & \textbf{0.0787↓} & \textbf{0.6738↓}  & \textbf{0.1356↓} & \textbf{0.0327↓} \\
            \hline
            \end{tabular}
        \end{adjustbox}
        \caption{Comparison of Depth Estimation Metrics on KITTI and Custom Orchard datasets. Up arrows (↑) indicate improvement for higher-better metrics, and down arrows (↓) indicate improvement for lower-better metrics.}
        \label{tab:comparison_metrics}
    \end{table*}

    \begin{figure*}[ht]
        \centering
        \includegraphics[width=0.87\linewidth]{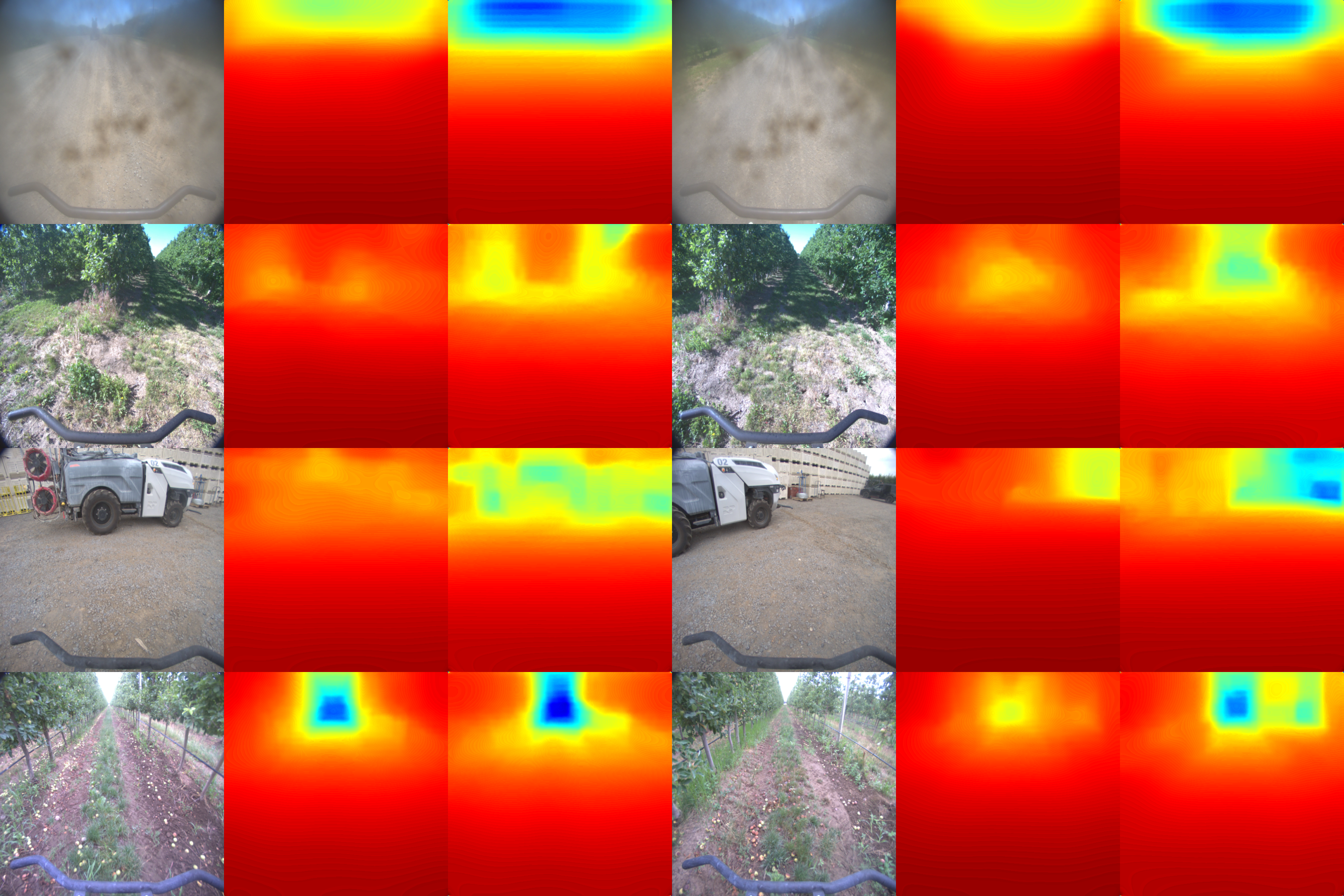}
        \caption{Visual result of metric depth estimation. The images are in RGB, Predict SiLog Loss, and Predict Consistency Loss order. We can observe that the depth estimation results represent a difference. The depth predicted by the model train with consistency loss will make the depth between points smoother.}
        \label{fig:result display}
    \end{figure*}
    
\section{Experiments}
    \subsection{Implementation}
        Considering the balance between efficiency and performance, in our experiment we implemented ViT-Base \cite{dosovitskiy2020vit} as the encoder and DPT \cite{dpt} as the decoder, and trained it on a single Nvidia 4090Ti GPU with 24GB of memory.
        
        To enable our model to predict cross-data metric depths across multiple datasets, we train our model on the KITTI dataset with a stereo-generated dense depth map and a custom dataset collected simultaneously in orchards. However, as our custom data has only sparse depth information projected from LiDAR points, we apply a sampler on top of the data loader to load different data categories. The sampler returns the sparse and dense depth map when loading the KITTI dataset \cite{kittdepthbenchmard,kitti}, and loads sparse data plus an empty tensor as a placeholder for dense depth when loading the custom dataset. We feed the neural network images by resizing them to scale \(\in\) [0.85,1.15] and then randomly cropping them to size [$518 \times 518$].

        We start our training with a learning rate of $1 \times 10^{-5}$ and batch size = 8, using AdamW as our model optimizer with $\beta \in [0.9, 0.999]$ and weight decay 0.01. We use cosine annealing as our learning rate scheduler to prevent overfitting, with parameters $T_{\text{max}} = 35$ and $ETA_{\text{min}} = 1 \times 10^{-8}$. We also apply the gradient of a norm clip-on model to prevent gradient explosion with max\_norm = 1.0 and norm\_type = 2.
        
    \subsection{Evaluation}
        In order to evaluate the performance of our model, we have chosen to follow the most common criteria used in depth estimation benchmarks. In particular, we used the accuracy under threshold \(\delta_{i} < 1.25^{i}\), \(i \in (1,3)\), absolute relative error (AbsRel), root mean square error (RMSE), log root mean square error (RMSE\_log) and log10 error (log10) to validate our dataset.

        In particular, the accuracy below the threshold can be represented as

        \begin{equation}
            \delta_i =  \max\left(\frac{D_i}{D_{i, gt}}, \frac{D_{i, gt}}{D_i}\right) < 1.25^i 
        \end{equation}

       absolute relative error:
        \begin{equation}
            \text{AbsRel} = \frac{1}{N} \sum_{i=1}^{N} \frac{|D_i - D_{i, gt}|}{D_{i, gt}}
        \end{equation}

        root mean square error:
        \begin{equation}
            \text{RMSE} = \sqrt{\frac{1}{N} \sum_{i=1}^{N} (D_i - D_{i, gt})^2}
        \end{equation}

        root mean square error in the log:
        \begin{equation}
            \text{RMSE\_log} = \sqrt{\frac{1}{N} \sum_{i=1}^{N} \left(\log D_i - \log D_{i, gt}\right)^2}
        \end{equation}

        log10 error:
        \begin{equation}
            \text{log10} = \frac{1}{N} \sum_{i=1}^{N} \left| \log_{10} D_i - \log_{10} D_{i, gt} \right|
        \end{equation}

    \subsection{Result}
        To evaluate the validation and efficiency of the Dense-Sparse Consistency Loss, we train our model with SiLog and Dense-Sparse Consistency Loss, respectively. The result is shown in table \ref{tab:comparison_metrics} and figure \ref{fig:result display}.  

        Table \ref{tab:comparison_metrics} shows that the loss of consistency exceeds critical criteria such as $\delta_{1}$, absolute relative error, and root mean square error. Especially for the RMSE, all three validation data sets show a massive improvement. The reason for the improvement in RMSE is that the dense depth map generated by the stereo camera contains more spatial and depth information than the projected point clouds, which helps the model to understand the distribution between objects and the spatial continuity in scene space.

        Meanwhile, the $\delta_1$ and AbsRel in the validation dataset have been significantly improved in the dense depth map dataset by dense-sparse consistency supervision. In contrast, the difference in $\delta_1$ is not massive in the sparse points. This result shows that training a depth prediction model using only the sparse point cloud is insufficient. The sparse point validation dataset cannot correctly validate the depth estimation value in the region without depth data. Therefore, it performs poorly in these pixels, significantly reducing the  $\delta_1$ in the dense validation set. On the other hand, the model training with sparse depth consistency precisely predicts depth in the whole image without degrading the depth prediction in the sparse validation data.

        Furthermore, the experimental results show that the dense depth information from a more thorough dataset in cross-dataset training can help a dataset containing only sparse points to better understand the depth distribution in the missing point areas. In the case of dense-sparse consistency supervision loss, the gradient wrongly calculated in the gap between sparse points will be fixed during the dense-sparse consistency validation. Therefore, it helps sparse points to align the depth distribution transit between neighboring points in different objects in multiple scenes.
    
\section{Discussion}
    Our custom datasets are collected in the apple orchard, which contains only limited types of crops that are planted with posts, but it could easily be applied to unseen crop types. Our model can accurately estimate depth in most scenes with different crop types if the encoder can extract the correct feature binding with the depth correspondence during decoding. This is because the ViT with DPT architecture has been shown to generate robust features for zero-shot capability, which has been widely used by models such as Depth Anything, Depth AnythingV2, etc. The training of significant data is critical to achieve robust performance for this type of model. Therefore, our model can be used directly in most orchard environments, only a few edge cases may require some fine-tuning to achieve better performance.
    
    The experiment was extended to ascertain the discrepancy between the stereo map of the KITTI dataset and the ground truth points, and to validate the measurement disparity between the rectified projected depth from the LiDAR ground truth and the depth map generated by the stereo pair from image02 (left camera) and image03 (right camera) with intrinsic calibration and baseline.  The results are presented in Table 1, which compares the valid points (depth greater than zero) of the LiDAR project depth and stereo depth.
    
    \begin{table}[ht]
        \centering
        \begin{tabular}{c|ccc}
        \hline
             d \textgreater 0 &L1&MSE&Variance\\
        \hline
             Minimum& 0.0704 & 0.0542 & 0.0538\\
             Maximum& 16.07359 & 611.5913 & 439.8481\\
             Average& 1.9072 & 45.4513 & 42.6781\\
        \hline
        \end{tabular}
        \caption{Dense sparse ground truth disparity in meters}
        \label{tab:GT_Disparity_one}
    \end{table}

    We suspect that the transparent objects and distance points are the cause of the significant outliers. To validate this, we constrained the depth value from the sparse and dense depth maps to less than 80 and 120 meters, respectively, and excluded points from transparent objects; then, the disparity between the depth from the dense depth map and sparse points was significantly reduced. This finding is consistent with our experimental result in the previous section when we designed the sparse-dense consistency loss function and implemented MSE to supervise disparity, which performs better than L1 and gradient cosine similarity because it massively punishes the mismatching outlier points. We suspect that the application of dense-sparse consistency loss with a depth range below 80 meters may improve depth estimation, which we will ensure in our further work.

    Compared to the KITTI dataset from DepthAnything or Metric3D, the performance of our model is slightly below their best result. Apart from the hardware disparity and backbone difference (ViT-large/giant vs. ViT-Base), the significant difference in resolution ratio, depth distribution, and training range of the datasets may be responsible for this. We suspect that synchronizing the depth range of the KITTI dataset with that of the orchard dataset may result in better performance. However, this will result in a degradation of the KITTI dataset, which we can demonstrate in our next work.
    
\section{Conclusions}
    In conclusion, this paper proposes a novel model to overcome the current constraints in metric depth estimation in the orchard/vineyard setting. Moreover, we have proposed a novel approach for leveraging a more comprehensive dataset that encompasses dense and sparse depth ground truth, aiming to facilitate the learning of transit distributions between valid points in a less comprehensive dataset that contains only sparse LiDAR points. The results of the experiment presented in this paper demonstrate that averaging the dense depth map from another dataset can enhance the robustness of our less comprehensive dataset in terms of regression in the spatial domain, even in the absence of dense depth information. This approach has achieved state-of-the-art results on both public and custom datasets.

\bibliography{publications}
\bibliographystyle{named}
\end{document}